\documentclass[runningheads]{llncs}

\usepackage{eccv}

\usepackage{eccvabbrv}

\usepackage{graphicx}
\usepackage{booktabs}
\usepackage[accsupp]{axessibility}  
\usepackage{amsmath}
\usepackage{capt-of}

\usepackage{hyperref}
\usepackage[table]{xcolor}
\definecolor{qwenrow}{gray}{0.92}
\definecolor{unetrow}{gray}{0.85}
\definecolor{oraclerow}{RGB}{225,235,245}
\definecolor{apgrow}{RGB}{255, 255, 224}

\usepackage{orcidlink}

\usepackage{placeins}

\begin{document}

\title{Adapting Open-Weight MLLMs to Generate Point Prompts for
Electron Microscopy Segmentation} 

\titlerunning{Open-Weight MLLM Point Prompts for EM Segmentation}

\author{Samia Mohinta\inst{1,2}\orcidlink{0000-0002-6675-5006}\thanks{Corresponding author email: sm2667@cam.ac.uk} \and
Albert Cardona \inst{2,1}\orcidlink{0000-0003-4941-6536}
}

\authorrunning{S. Mohinta, A. Cardona}

\institute{University of Cambridge, Cambridge, UK \and
MRC Laboratory of Molecular Biology, Cambridge, UK
}

\maketitle

\begin{abstract}
Promptable models such as microSAM segment electron microscopy (EM) images
from point prompts, but automation requires generating prompts without user
input. We ask whether open-weight multimodal large language models (MLLMs)
can generate them from natural-language requests by returning coordinates to a
frozen segmenter. To that end, we convert masks from three mitochondria
datasets into training examples, pairing images and instructions with centroid
coordinates, then train LoRA adapters while freezing the MLLM backbone and
microSAM. We find that Qwen3-VL reaches segmentation AP$_{50}$ $0.736$
after supervised fine-tuning and reward optimization, up from $0.247$ without
adaptation, while automatic prompt generation (APG) achieves $0.773$. In addition,
two other MLLMs improve, reaching or exceeding APG. When compared with a
supervised centroid-heatmap detector that reaches AP$_{50}$ $0.904$ for this
mitochondria task, Qwen3-VL more closely matches the annotated point set and
instance counts. Moreover, training on two public datasets transfers to an
unseen third, while training on all three transfers to an independent EM volume.
Robustness tests show stable performance under unseen formulations of the
natural-language request, while the coordinates can be reused by a second
segmenter. To our knowledge, this is the first feasibility study of open-weight
MLLMs as EM point generators, providing an inspectable, language-directed
link between localization and mask decoding.

\end{abstract}

\section{Introduction}
 
Promptable segmentation models have become increasingly common in biological
image analysis, including electron microscopy (EM). Examples include
Segment Anything Model (SAM)-based~\cite{kirillov2023sam} segmenters such
as microSAM~\cite{archit2025microsam} and CellSAM~\cite{marks2025cellsam},
along with text-guided models such as
BiomedParse~\cite{zhao2024biomedparse}. Depending on the model, prompts may
take the form of points, boxes, or text phrases~\cite{kirillov2023sam,archit2025microsam,zhao2024biomedparse}, specifying the target without
requiring a hand-traced boundary. In
practice, users supply these prompts manually, for example by placing points
on the objects of interest~\cite{archit2025microsam}. But fully automatic operation requires a source
that predicts where those points should be placed. The point source therefore
determines if manual clicks are required.
 
\begin{figure*}[ht!]
\centering
\includegraphics[width=\textwidth]{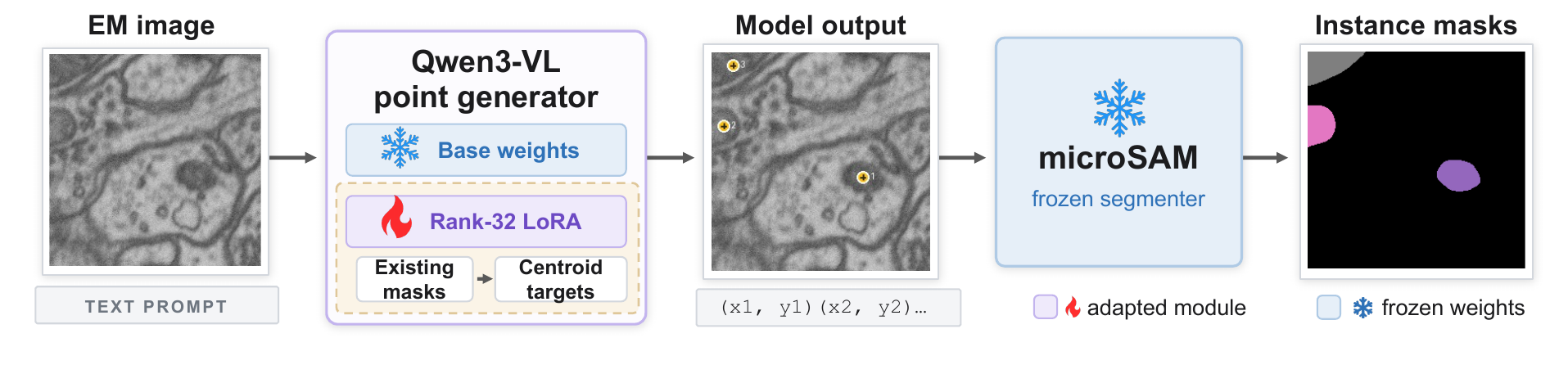}
\caption{\textbf{Mask-derived supervision turns an open-weight MLLM into an inspectable point generator for a frozen segmenter.}
Segmentation masks are reduced to centroid targets that train a low-rank adapter
on Qwen3-VL~\cite{bai2025qwen3vl}. Given an EM crop and a natural-language
request, the adapted model emits coordinates that the frozen microSAM decodes
into instance masks.}
\label{fig:overview}
\end{figure*}

Automatic prompt generation (APG) in microSAM is one such source. It derives
candidate points from microSAM's own segmentation-decoder predictions,
decodes masks from those points, and filters overlapping masks without
retraining~\cite{archit2026apg}. This provides an effective automatic mode,
although the localization remains determined by the decoder's predictions and
is not conditioned on a natural-language request. We reconsider this
localization stage.

Multimodal large language models (MLLMs) trained for visual grounding offer
a different interface. Given an image and a target expressed in natural
language, they can return the corresponding image locations as explicit
coordinates in their text output~\cite{peng2023kosmos2,deitke2024molmo}.
Localization then becomes an intermediate prediction that can be evaluated
independently before mask decoding. Because the coordinates are expressed in
image space, the same output can also be supplied to another point-prompted
segmentation model.

We assess whether this interface is viable using mitochondria in EM as a
test bed, primarily because mitochondria segmentation is a well-established
task that dedicated models already handle well~\cite{conrad2023mitonet}
and for which multiple public benchmark datasets are available, including
Lucchi++~\cite{lucchi2012,casser2020lucchiplus},
VNC~\cite{gerhard2013vnc}, and MitoEM-H~\cite{wei2020mitoem}. These
datasets supply the annotated masks we need and let us vary the EM condition
while keeping the target fixed. However, transferring this interface to EM may still fail,
since MLLM pointing has been demonstrated mainly on natural
images~\cite{cheng2025pointarena,deitke2024molmo}. EM images contain densely
packed ultrastructure with limited local contrast~\cite{liu2022semisup},
mitochondrial appearance varies across species, tissues, and acquisition
conditions~\cite{conrad2023mitonet,wei2020mitoem}, and lightweight
adaptation of vision foundation models does not always close the domain gap
in EM~\cite{fusterbarcel2026visionfoundationmodelsfoundational}. We
therefore ask first whether open MLLMs can localize mitochondria in EM, then
whether parameter-efficient adaptation makes them usable as point sources,
and finally whether the learned localization transfers to unseen conditions.

Dedicated pointing datasets provide supervision for coordinate prediction
in natural images and
robotics~\cite{deitke2024molmo,yuan2024robopoint}. By contrast, the three
EM benchmarks used here provide segmentation masks without dedicated point
targets. We assign one centroid target to each annotated mitochondrion in
these benchmarks and express these targets as coordinate instructions. We
then adapt open-weight MLLMs with Low-Rank Adaptation
(LoRA)~\cite{hu2022lora}, training only the LoRA parameters and leaving the
pretrained MLLM backbone and microSAM unchanged
(Figure~\ref{fig:overview}).

Adaptation proceeds in two stages. Supervised fine-tuning (SFT) trains the
model to localize mitochondria and serialize the predicted locations in the
required coordinate format. Group Relative Policy Optimization
(GRPO)~\cite{shao2024deepseekmath} then optimizes a reward over response
validity, point quality, and the overlap of masks decoded from the predicted
points. Qwen3-VL is our primary model because its grounding interface
already represents image locations as relative coordinates in its text
output~\cite{bai2025qwen3vl}, matching the representation our pipeline
requires. We apply the same mask-derived SFT procedure to
InternVL3.5~\cite{wang2025internvl35} and Molmo2, which is trained for
point-based grounding~\cite{clark2026molmo2}, to test whether adaptation
extends across MLLM backbones. Transfer is assessed by training on two of
the three benchmarks and evaluating on the excluded one, and by applying the
adapter trained on all three benchmarks to an independent EM volume. To place the MLLM point-generation interface in context, we compare the
adapted MLLMs with APG and a lightweight centroid-heatmap detector trained
on the same crops and centroid targets. A U-Net trained on the same masks
serves as a supervised reference.

\noindent
\textbf{Contributions.}
\begin{itemize}

\item To our knowledge, the first feasibility study of open-weight MLLMs as
executable point-prompt generators for EM segmentation, with point
supervision derived from existing segmentation masks without additional manual
point annotation.

\item A staged analysis of zero-shot inference, SFT, and GRPO on Qwen3-VL,
with the same mask-derived SFT procedure tested on InternVL3.5 and Molmo2
and evaluated against APG and a supervised centroid-heatmap point source.

\item Leave-one-dataset-out transfer experiments on public EM datasets,
together with evaluation on an independent EM volume and tests of unseen
instruction phrasings and a second segmentation backend.

\end{itemize}

\section{Related Work}
\label{sec:related}

\textbf{Promptable microscopy and automatic prompt sources.}
Promptable segmentation has been adapted to several biomedical imaging
settings. microSAM adapts the Segment Anything Model
(SAM)~\cite{kirillov2023sam} for interactive and automatic segmentation
in light and electron microscopy~\cite{archit2025microsam}. CellSAM pairs
SAM with a learned detector that supplies box
prompts~\cite{marks2025cellsam}, while PathoSAM supports automatic and
interactive nucleus segmentation in
histopathology~\cite{griebel2025pathosam}. SAM 3 further allows
concept-conditioned segmentation from noun phrases and image
exemplars~\cite{carion2025sam3}. Across these systems, the prompt specifies
what or where to segment; for automatic use, the remaining question is how
that prompt is obtained.

Different systems obtain automatic prompts in different ways. SAM's
automatic mask generator evaluates a dense grid of points and filters the
resulting masks~\cite{kirillov2023sam}. CellSAM predicts box prompts with
an object detector~\cite{marks2025cellsam}, while APG derives candidate
points from microSAM's learned instance predictions and filters the decoded
masks~\cite{archit2025microsam,archit2026apg}. UN-SAM learns
self-generated mask hints within an automatic nuclei-segmentation
system~\cite{chen2024unsam}. In these approaches, localization remains
within the segmentation system or an associated detector. We examine a different division of labor: a separate MLLM generates point
prompts from natural-language requests, while the microscopy segmenter
remains fixed.

\textbf{Language-conditioned segmentation through latent representations
or explicit geometry.}
One line of work connects language to masks through learned latent
representations. BiomedParse conditions a biomedical segmentation decoder
on text~\cite{zhao2024biomedparse}. LISA decodes the hidden representation
of a special segmentation token into a mask~\cite{lai2024lisa}, while
PixelLM uses segmentation-codebook tokens and a dedicated pixel
decoder~\cite{ren2024pixellm}. Although their architectures differ, these
methods produce masks from learned representations rather than returning
the selected locations as a discrete coordinate list.

A second line of work makes the spatial output explicit. Kosmos-2~\cite{peng2023kosmos2} and
Molmo~\cite{deitke2024molmo} demonstrate that multimodal models can express visual grounding as coordinates in their text outputs. SAM4MLLM~\cite{sam4mllm2024} uses an MLLM to find point prompts for SAM in referring-expression
segmentation. Seg-Zero~\cite{liu2025segzero} produces positional prompts for
a separate segmentation model, and Seg-R1~\cite{segr12025} generates
point and box prompts for SAM2 using
GRPO. Think2Seg-RS~\cite{think2segrs2025} trains a vision-language model to
control a frozen SAM through structured geometric prompts in remote
sensing. GenSeg-R1~\cite{gensegr12026} similarly trains Qwen3-VL to
emit a bounding box and interior keypoints for a frozen SAM2
backend. These studies establish MLLM-generated geometric prompts in natural images
and remote sensing, but leave open whether point generation can be learned
for EM, where the visual domain and available supervision differ
substantially.

\textbf{Coordinate-generating models in biomedical imaging.}
Several recent biomedical systems provide closer comparisons.
SmartPath-R1~\cite{smartpathr12025} generates bounding-box coordinates that are passed to MedSAM
for pathology segmentation, while IBISAgent~\cite{ibisagent2026} produces
iterative text-based click actions and invokes biomedical segmentation
tools. NuNext~\cite{nunext2026} formulates nucleus detection in
histopathology as autoregressive centroid
prediction. These methods provide biomedical precedents for explicit coordinate
prediction, but use different imaging modalities and prompt interfaces.

uLLSAM~\cite{ulsam2025} is the closest microscopy-specific precedent. It
injects MLLM-derived vision-language knowledge into SAM through an
alignment module and evaluates the resulting model across microscopy
datasets that include EM. It does not, however, train the MLLM to return
point coordinates. Our setting keeps the pretrained MLLM backbone and
segmentation model fixed, adapting only a lightweight module to emit the
points passed between them. MicroVQA also provides a complementary example
of MLLM evaluation in microscopy through question
answering~\cite{burgess2025microvqa}; we evaluate the predicted locations
directly and through the instance masks they produce.

\textbf{Parameter-efficient adaptation and task-level optimization.}
LoRA introduces trainable low-rank updates into selected linear
transformations while leaving the pretrained weights
fixed~\cite{hu2022lora}. We use LoRA to learn from point targets derived
from existing segmentation masks. Beyond SFT, GRPO enables optimization against
structured task rewards without training a separate value
model~\cite{shao2024deepseekmath}. Reinforcement learning has also been
used for positional prompts in Seg-Zero, Seg-R1, and
GenSeg-R1~\cite{liu2025segzero,segr12025,gensegr12026}, for iterative
biomedical click actions in IBISAgent~\cite{ibisagent2026}, and for
centroid prediction in NuNext~\cite{nunext2026}. Through our experiments,
we measure how much each adaptation stage contributes to point localization
in EM.

\section{Method}
\label{sec:method}

\subsection{Datasets}
\label{subsec:datasets}

We use three public EM mitochondria benchmarks: Lucchi++
~\cite{lucchi2012,casser2020lucchiplus}, Drosophila
VNC~\cite{gerhard2013vnc}, and MitoEM-H~\cite{wei2020mitoem}. Within each
dataset, the two-dimensional source slices are split into $80\%$ training,
$10\%$ validation, and $10\%$ test sets using seed $42$. Slices in each partition are then tiled into $256\times256$ crops with
stride $128$. Train, validation, and test partitions are therefore
source-slice disjoint in all three datasets, including VNC, but not
volume-disjoint; adjacent sections may occur across partitions. The test
sets contain $342$, $79$, and $332$ crops, respectively.

The \emph{combined} setting joins the three training sets. ``All data''
denotes evaluation over the union of the three test sets, comprising $753$
crops.

\subsection{Point generation and mask decoding}
\label{subsec:interface}

Given an EM crop $I$ and an instruction $q$, an MLLM produces a text
response $s$ from which we parse valid normalized coordinates,
\begin{equation}
s=f_\theta(I,q), \qquad
P=\operatorname{parse}(s)=\{(x_i,y_i)\}_{i=1}^{\hat N_{\mathrm{pt}}},
\qquad \hat N_{\mathrm{pt}}=|P|,
\end{equation}
where $x_i,y_i\in[0,1]$, the origin is at the top left, and $x$ and $y$
increase rightward and downward. Malformed and out-of-range point tokens
are discarded. The response also states a textual count, but the parsed
count $\hat N_{\mathrm{pt}}$ determines the GRPO count reward and
point-count mean absolute error (MAE)~\cite{Willmott2005}.

Each parsed coordinate is passed to frozen microSAM as a positive prompt
and yields one selected mask. A mask is suppressed when its intersection
over union (IoU) with an already accepted mask is at least $0.80$;
remaining overlap pixels are assigned to the first accepted instance.

Point supervision is derived from the existing segmentation labels.
Stored instance identifiers are used when available, while semantic masks
such as VNC are separated by connected components. Each instance is
represented by its normalized arithmetic centroid. Crops without
mitochondria are retained with probability $15\%$. The target response
states the count in words and contains one point token per instance. For
cross-entropy SFT, points are serialized in the deterministic order
provided by the labels: ascending instance identifier for instance maps
and connected-component scan order for semantic masks. Serialization and parser details are provided in Supplementary
Sections~\ref{app:implementation} and~\ref{app:reward}.

\subsection{Adapting the point generator}
\label{subsec:adaptation}

Standard evaluation uses the instruction ``Identify all mitochondria in this
electron microscopy image.''

\paragraph{Zero-shot.}
All three MLLMs are evaluated with the standard instruction and no EM
adaptation. Model-specific preprocessing and decoding settings are given in
Supplementary Section~\ref{app:implementation}.

\paragraph{Supervised fine-tuning.}
We apply LoRA~\cite{hu2022lora} while keeping the pretrained MLLM weights
fixed. Our primary Qwen3-VL setting uses rank $32$, $\alpha=64$, dropout
$0.05$, learning rate $2\times10^{-5}$, effective batch size $8$, and three
SFT epochs on the combined training set with seed $42$. The Qwen3-VL adapter
updates approximately 1\% of the model parameters. We also train separate dataset-specific Qwen3-VL adapters on Lucchi++,
VNC, and MitoEM-H using seed $42$ and the same three-epoch SFT schedule,
followed by one GRPO epoch.

For the cross-backbone comparison, Qwen3-VL, InternVL3.5, and Molmo2 are each
trained for one SFT epoch on the same $5{,}957$-crop combined training set
with the same effective batch size. We use short fixed SFT schedules, as is
common in multimodal instruction tuning~\cite{liu2023llava,liu2024llava15}.

Training cycles through $35$ mitochondria instructions. A separate robustness
test evaluates five additional phrasings absent from the training pool. We
also test three morphology-only phrasings that omit mitochondrial terminology
at inference using the unadapted Qwen3-VL model and its existing SFT
checkpoint.

We run detailed ablations on Qwen3-VL because it has the lowest observed
SFT memory use and training time per epoch among the three models: $22.3$~GB
and approximately $1$ hour, versus $25.8$~GB and $2.2$ hours for InternVL3.5
and $47.7$~GB and $3$ hours for Molmo2 on an NVIDIA H200. A data-scaling
experiment trains Qwen3-VL on nested subsets of the Lucchi++ training slices,
with each adapter evaluated on the same fixed Lucchi++ test set. An
adapter-rank ablation evaluates $r\in\{4,8,16,32,64\}$. Full training and decoding
settings are reported in Supplementary Section~\ref{app:implementation}.

\paragraph{Reward optimization.}
GRPO~\cite{shao2024deepseekmath} continues from Qwen3-VL SFT checkpoints for
one epoch, taking approximately $16$ hours on the H200, with
learning rate $5\times10^{-7}$ and four sampled responses per crop. The reward
combines format validity, parsed point-count accuracy, centroid-distance
matching, the fraction of predicted points on foreground, and
Hungarian-matched mask IoU~\cite{kuhn1955hungarian}. A multiplicative penalty
discourages degenerate coordinate outputs. microSAM is used only to compute
the mask-IoU reward and is not differentiated through. Reward equations and
weights are provided in Supplementary Section~\ref{app:reward}. The combined
SFT and SFT+GRPO comparison is repeated with seeds $7$, $42$, and $123$.

\subsection{Transfer experiments}
\label{subsec:transfer_setup}

For each leave-one-dataset-out condition, a separate one-epoch Qwen3-VL
adapter and a centroid-heatmap detector are trained on two public
datasets and evaluated on the unchanged test split of the excluded third.
Moreover, a Qwen3-VL adapter is trained for three epochs on VNC and
MitoEM-H and evaluated on the same held-out Lucchi++ test set.

Because VNC has the fewest training examples, we use its $433$-crop training
set as a matched training-set size to examine the effect of data composition.
We compare three one-epoch SFT conditions containing $433$ unique crops each:
VNC-only, a mixture of all three datasets, and Lucchi++ and MitoEM-H without
VNC. Each condition is trained with seeds $7$, $42$, and $123$ and evaluated
on the same $79$-crop VNC test set.

We additionally evaluate the unchanged combined three-epoch SFT adapter on a
separate in-house larval-fly EM volume that is absent from adaptation. The
reported test set is a fixed $131$-crop subset, corresponding to $1\%$ of the
$13{,}104$ generated crops and selected before model evaluation. The
pretrained microSAM checkpoint remains frozen in all transfer experiments.

\subsection{Segmentation backends and comparisons}
\label{subsec:baselines}

All main point-source comparisons use the frozen microSAM
\texttt{vit\_b\_em\_\allowbreak organelles}
checkpoint~\cite{archit2025microsam}. For backend controls, we decode saved Qwen3-VL GRPO points with
\texttt{vit\_l\_em\_\allowbreak organelles} and saved Qwen3-VL SFT
points with SAM 3~\cite{carion2025sam3}.

Our internal automatic point-source comparison is microSAM's APG~\cite{archit2026apg}. We also replace the learned
generator with centroids from the ground-truth instances, denoted
\emph{oracle centroids}, to measure the fixed microSAM pipeline under
ground-truth point placement.

For a supervised point-source comparison, we train a
$488{,}417$-parameter U-Net-style centroid-heatmap detector
~\cite{ronneberger2015unet_seg} on the same crops and centroid targets as
the MLLMs. Local heatmap maxima are filtered by distance-based
non-maximum suppression and passed to the same frozen microSAM backend.
We train combined and leave-one-dataset-out variants; details are given in Supplementary
Section~\ref{app:heatmap}.

For dense-mask context, we train a combined
$7{,}851{,}969$-parameter U-Net~\cite{ronneberger2015unet_seg} on the
three public training sets using binary cross-entropy and Dice
losses~\cite{milletari2016vnet}. At inference, foreground probabilities
are thresholded at $0.5$, connected components form instances, and each
instance is scored by its mean foreground probability. Full architecture
and optimization details are provided in Supplementary
Section~\ref{app:unet}.

\subsection{Evaluation}
\label{subsec:eval}

We evaluate point generation before mask decoding and segmentation after
the prompts are executed. Dataset-specific and ``All data'' results are
arithmetic means of crop-level metrics.

\paragraph{Point metrics.}
Let $P=\{p_i\}$ be the predicted points and $G=\{g_j\}$ the ground-truth
centroids. Within each crop, we compute a one-to-one Hungarian assignment
~\cite{kuhn1955hungarian} using normalized Euclidean distance. A matched
pair is a true positive when its distance is at most $\tau=0.05$;
unmatched predictions and targets are false positives and false negatives.
We compute precision, recall, and their harmonic mean for each crop and
report the mean crop-level point F1. If both sets are empty, the crop
scores $1$; if only one is empty, it scores $0$.

Point-in-mask precision is the fraction of predicted points inside any
ground-truth instance. Object recall is the fraction of ground-truth
instances containing at least one predicted point. Both are computed per
crop and then averaged. A diagnostic point F1 at $\tau=0.10$ and point-count MAE, the mean
absolute difference between $\hat N_{\mathrm{pt}}$ and the ground-truth
instance count, are reported in Supplementary
Section~\ref{app:point_mask}.

\paragraph{Segmentation metrics.}
We report foreground Dice, average precision, mean segmentation accuracy
(mSA), and final instance-count MAE. For point-prompted methods, masks are
ranked by microSAM's predicted-IoU confidence; U-Net instances are ranked
by mean foreground probability. Predictions are greedily matched in score
order to unmatched ground-truth instances by mask IoU. Per-crop average
precision is computed with $101$-point interpolation and averaged across
crops. Following COCO~\cite{lin2014coco}, we report AP$_{50}$,
AP$_{75}$, and mAP over
$\mathcal{T}=\{0.50,0.55,\ldots,0.95\}$, with AP$_{50}$ as the primary
measure.

Mean segmentation accuracy averages
$\mathrm{TP}(t)/[\mathrm{TP}(t)+\mathrm{FP}(t)+\mathrm{FN}(t)]$ over the
same thresholds, using Hungarian mask matching at each threshold
~\cite{caicedo2019nucleus,archit2026apg}. Final instance-count MAE is
computed after mask decoding and duplicate suppression, so it can differ
from point-count MAE when several prompts produce one retained instance.

\section{Results and Discussion}
\label{sec:results}

\subsection{Mask-derived adaptation teaches the model where to look}
\label{subsec:points}

\begin{figure*}[ht!]
\centering
\includegraphics[width=\textwidth]{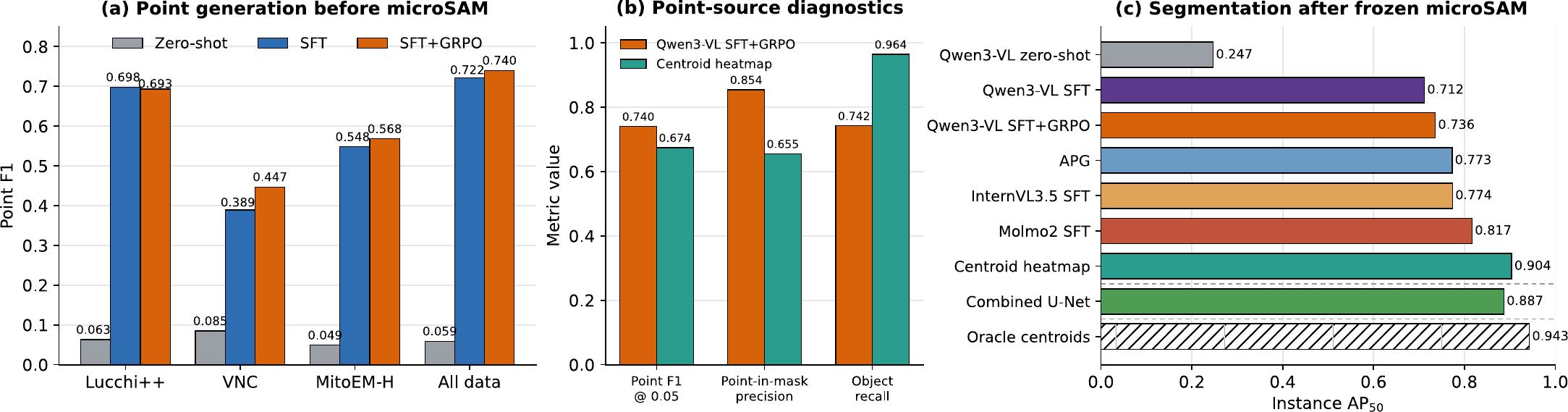}
\caption{\textbf{Point localization, point-source diagnostics, and
downstream segmentation.}
\textbf{(a)} Qwen3-VL point F1 at $\tau=0.05$. Lucchi++, VNC, and
MitoEM-H show evaluation on the individual datasets; the SFT and
SFT+GRPO bars use adapters trained separately on each dataset. ``All data'' uses the combined adapter.
\textbf{(b)} Point-source diagnostics for Qwen3-VL SFT+GRPO and
the supervised centroid-heatmap detector on ``All data''.
\textbf{(c)} Instance AP$_{50}$ over the same test set. Point sources are decoded
by frozen microSAM; the U-Net predicts masks directly. Qwen3-VL SFT and
SFT+GRPO use the seed-42 run with three SFT epochs, whereas InternVL3.5
and Molmo2 use one. Figure~\ref{fig:panel}d gives the matched one-epoch
comparison. Oracle centroids prompt every annotated instance at its
ground-truth centroid.}
\label{fig:main}
\end{figure*}

Zero-shot Qwen3-VL produces parseable coordinates but localizes poorly,
reaching point F1 $0.059$ at $\tau=0.05$. SFT using the mask-derived
centroid targets raises point F1 to $0.722$, and GRPO increases it to
$0.740$ (Fig.~\ref{fig:main}a).

On the other hand, point F1 is $0.674$ for the supervised
centroid-heatmap detector and $0.740$ for Qwen3-VL SFT+GRPO. The heatmap
detector has higher object recall ($0.964$ versus $0.742$) but lower
point-in-mask precision ($0.655$ versus $0.854$;
Fig.~\ref{fig:main}b). It produces more proposals
(Supplementary Section~\ref{app:point_ap}; Fig.~\ref{fig:qual}), which improves object coverage
but also adds unmatched or off-target points. In contrast, Qwen3-VL produces
points that more often fall on target instances and match the annotated
centroids, but it misses more objects.

Performance differs across the three EM datasets
(Fig.~\ref{fig:main}a). After SFT, point F1 is highest on Lucchi++ and
lowest on VNC. VNC also has the smallest per-dataset training set, with
$433$ crops, so this comparison alone does not tell us whether the lower
performance comes from the EM condition or the amount of supervision (revisited in Fig.~\ref{fig:panel}b).
GRPO improves point F1 on VNC and MitoEM-H but not on Lucchi++. Dataset sizes and the corresponding single-dataset AP$_{50}$ results are
reported in Supplementary Sections~\ref{app:data_stats} and
~\ref{app:singledataset}.

\subsection{Adapted points support competitive segmentation}
\label{subsec:endtoend}

\begin{table}[ht!]
\centering
\small
\setlength{\tabcolsep}{3.5pt}
\caption{\textbf{End-to-end results on ``All data''.}
Point generators are decoded by frozen microSAM; the combined U-Net
predicts masks directly. Qwen3-VL SFT and SFT+GRPO use the primary
seed-42 run with three SFT epochs, whereas InternVL3.5 and Molmo2 use
one SFT epoch. Best MLLM values are in \textbf{bold}; best Qwen3-VL
values are \underline{underlined}.}
\label{tab:overall}
\begin{tabular}{@{}ll cccccc@{}}
\toprule
Prompt source & Training
& AP$_{50}$ $\uparrow$
& AP$_{75}$ $\uparrow$
& mAP $\uparrow$
& Dice $\uparrow$
& mSA $\uparrow$
& MAE $\downarrow$ \\
\midrule

\multicolumn{8}{@{}l}{\textit{MLLM point generators}} \\
Qwen3-VL    & zero-shot & 0.247 & 0.235 & 0.202 & 0.323 & 0.183 & 1.182 \\
Qwen3-VL    & SFT       & 0.712 & 0.659 & 0.570 & 0.776 & 0.538 & 0.606 \\
Qwen3-VL    & SFT+GRPO  & \underline{0.736} & \underline{0.678}
& \underline{0.585} & \underline{0.792}
& \underline{0.547} & \underline{0.564} \\
InternVL3.5 & SFT       & 0.774 & 0.714 & 0.613 & 0.812 & 0.574 & 0.535 \\
Molmo2      & SFT       & \textbf{0.817} & \textbf{0.747}
& \textbf{0.644} & \textbf{0.836}
& \textbf{0.611} & \textbf{0.461} \\

\midrule
\multicolumn{8}{@{}l}{\textit{Non-language point generators}} \\
Centroid heatmap & supervised
& 0.904 & 0.801 & 0.691 & 0.817 & 0.456 & 1.483 \\
APG              & none
& 0.773 & 0.694 & 0.591 & 0.808 & 0.542 & 0.704 \\

\midrule
\multicolumn{8}{@{}l}{\textit{Direct segmentation and oracle}} \\
Combined U-Net & supervised
& 0.887 & 0.824 & 0.772 & 0.899 & 0.679 & 0.526 \\
Oracle centroids & --
& 0.943 & 0.815 & 0.707 & 0.906 & 0.693 & 0.003 \\
\bottomrule
\end{tabular}
\end{table}

On all $753$ ``All data'' test crops, the primary combined Qwen3-VL run has
AP$_{50}$ $0.247$ without adaptation, $0.712$ after SFT, and $0.736$
after GRPO (Fig.~\ref{fig:main}c; Table~\ref{tab:overall}). APG gives
$0.773$ on the same test set. The performance of the combined Qwen3-VL model is also stratified by
source dataset in Supplementary Section~\ref{app:percondition}.

The same mask-derived SFT procedure also improves all three MLLM
backbones under a matched one-epoch schedule. AP$_{50}$ rises from
$0.247$ to $0.648$ for Qwen3-VL, from $0.412$ to $0.774$ for
InternVL3.5, and from $0.015$ to $0.817$ for Molmo2
(Fig.~\ref{fig:panel}d). As a result, InternVL3.5 is effectively level with APG
($0.774$ versus $0.773$), while Molmo2 exceeds it and gives the
strongest MLLM results across the metrics in Table~\ref{tab:overall}.

As also noted in Sec.~\ref{subsec:points}, the centroid-heatmap detector
produces more point proposals, increasing the chance that each mitochondrion
receives at least one prompt and giving it higher object recall, but also
producing more unmatched or off-target points. After mask decoding, this greater object coverage contributes to its
AP$_{50}$ of $0.904$, higher than the adapted MLLMs and APG. But its higher
AP$_{50}$ does not imply more accurate point localization: additional
proposals can recover objects after decoding even when they reduce point-set
and count agreement. Qwen3-VL SFT+GRPO, for comparison, has higher mSA
($0.547$ versus $0.456$) and lower instance-count MAE
($0.564$ versus $1.483$). Supplementary
Section~\ref{app:point_ap} examines this effect by varying heatmap peak
suppression and measuring the resulting changes in point F1, AP$_{50}$,
and instance count.

Using the full segmentation masks for training, the combined U-Net gives
AP$_{50}$ $0.887$, mAP $0.772$, and Dice $0.899$
(Table~\ref{tab:overall}). Oracle centroids give AP$_{50}$ $0.943$, but
AP$_{75}$ falls to $0.815$ and mAP to $0.707$. Thus, even with
ground-truth point placement, the decoded masks do not always match the
annotated boundaries closely at stricter IoU thresholds. Supplementary Section~\ref{app:point_ap} further examines how different
point locations within the same ground-truth instances affect the masks
decoded by microSAM.

\begin{figure*}[ht!]
\centering
\includegraphics[width=\textwidth]{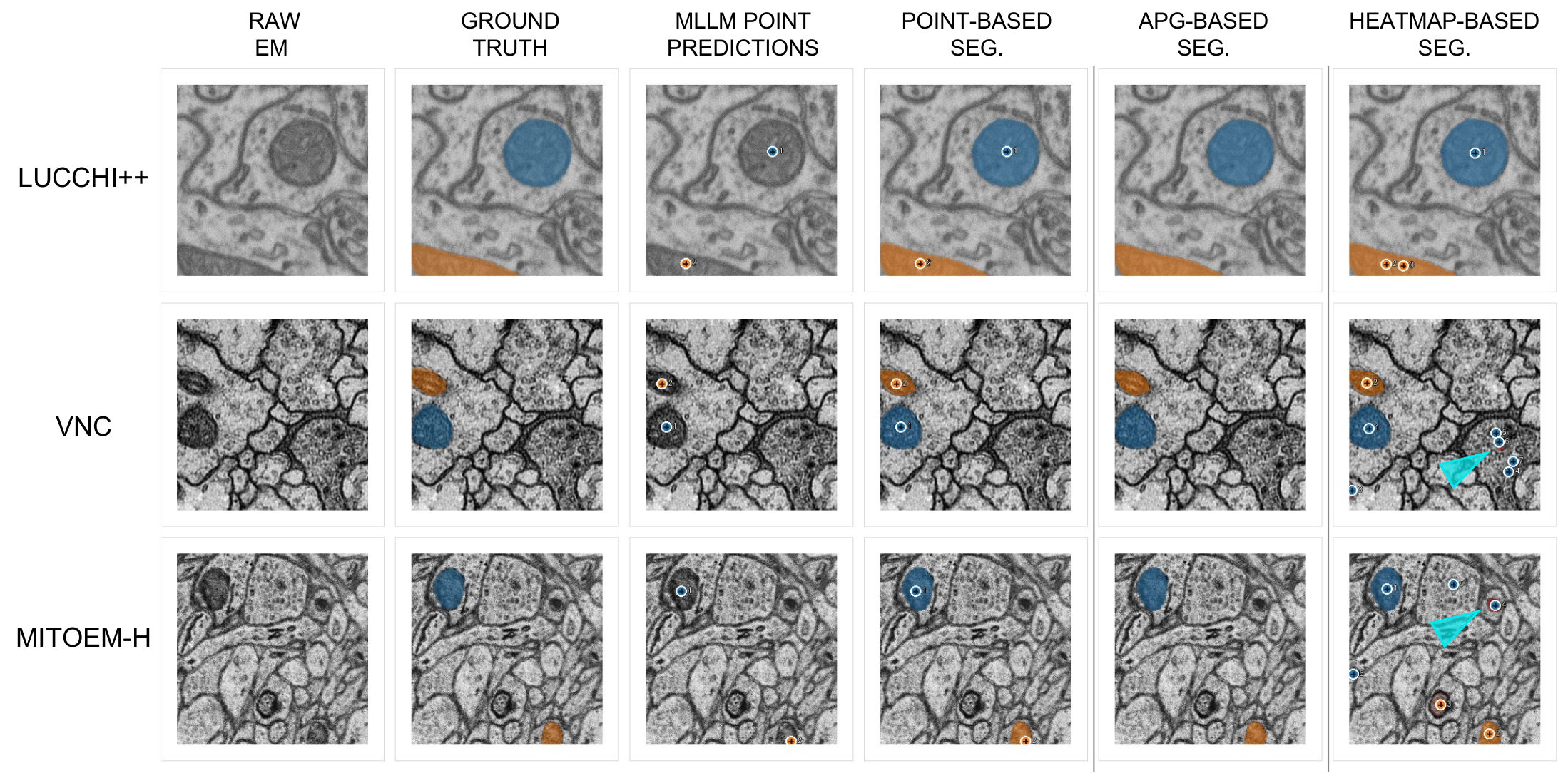}
\caption{\textbf{Point-to-mask hand-off on selected crops.}
Rows show Lucchi++, VNC, and MitoEM-H; columns show the raw EM image,
ground truth, Qwen3-VL SFT point predictions, masks decoded from those
points, APG masks, and masks decoded from centroid-heatmap points. Colors
distinguish instances; point markers indicate the prompts supplied to
microSAM. Cyan arrowheads indicate additional centroid-heatmap point
predictions that do not match annotated instances.}
\label{fig:qual}
\end{figure*}

Figure~\ref{fig:qual} shows how predicted coordinates become segmentation
masks in selected examples from all three datasets. For instances recovered by the different point sources,
the most visible differences occur around the decoded mask boundaries.

\subsection{Point generators transfer to held-out EM datasets and an independent volume}
\label{subsec:transfer}

\begin{figure*}[ht!]
\centering
\includegraphics[width=\textwidth]{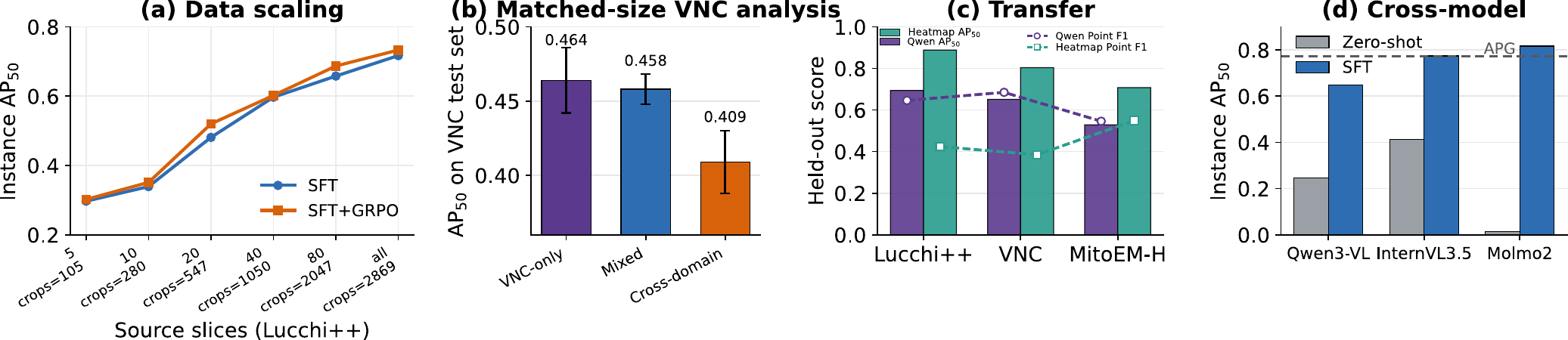}
\caption{\textbf{Training data, transfer, and cross-backbone adaptation.}
\textbf{(a)} Qwen3-VL SFT and SFT+GRPO AP$_{50}$ on the fixed Lucchi++
test set as training data increase.
\textbf{(b)} Matched-size VNC analysis using one-epoch Qwen3-VL SFT and
$433$ unique training crops from VNC only, a three-dataset mixture, or
Lucchi++ and MitoEM-H without VNC. Bars show mean AP$_{50}$ across three
seeds; error bars show standard deviation.
\textbf{(c)} Leave-one-dataset-out AP$_{50}$ (bars) and point F1
(lines) for separately trained Qwen3-VL and centroid-heatmap point
generators.
\textbf{(d)} Zero-shot and matched one-epoch SFT results across the three
MLLM backbones; the dashed line marks APG AP$_{50}$.}
\label{fig:panel}
\end{figure*}

When each public dataset is excluded from adaptation, both point generators
retain useful performance on its unchanged test set
(Fig.~\ref{fig:panel}c). The centroid-heatmap detector has higher
AP$_{50}$ on all three held-out datasets, with $0.888$ on Lucchi++,
$0.804$ on VNC, and $0.708$ on MitoEM-H. At the point level, Qwen3-VL
has higher point F1 on Lucchi++ ($0.646$ versus $0.424$) and VNC
($0.685$ versus $0.385$), while the two models are similar on MitoEM-H
($0.546$ and $0.549$). Thus, the centroid-heatmap detector retains stronger downstream
segmentation on the held-out datasets, while Qwen3-VL more closely matches
the annotated point set on two of the three.

On the  held-out Lucchi++ dataset, training the Qwen3-VL adapter for three epochs
on VNC and MitoEM-H raises AP$_{50}$ from $0.694$ after one epoch to
$0.712$. The VNC and MitoEM-H holdouts were not evaluated with the longer
schedule, so this result is only a training-duration check
(Supplementary Section~\ref{app:transfer_extension}).

VNC allows us to examine the effect of training-data composition at a fixed
budget. With $433$ training crops and one SFT epoch, VNC-only gives mean
AP$_{50}$ $0.464\pm0.022$, the three-dataset mixture gives
$0.458\pm0.010$, and training only on Lucchi++ and MitoEM-H gives
$0.409\pm0.021$ across three seeds (Fig.~\ref{fig:panel}b). Mixing
conditions therefore does not increase mean VNC performance when the
training-set size is fixed. In contrast, the leave-one-dataset-out model
trained on all $5{,}524$ Lucchi++ and MitoEM-H crops reaches
AP$_{50}$ $0.651$ without seeing VNC during adaptation. This shows that
localization learned from the other EM datasets can transfer to VNC when
substantially more cross-domain training data are available. Because one
epoch over the larger dataset also involves more optimizer updates, this
comparison does not separate the effects of additional examples from
additional training steps.

\begin{figure*}[t]
\centering
\includegraphics[width=\textwidth]{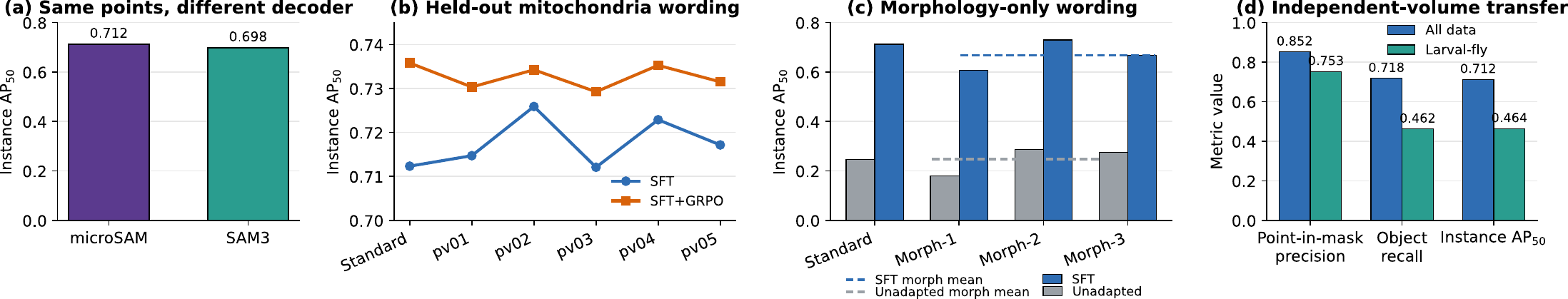}
\caption{\textbf{Decoder portability, instruction sensitivity, and
independent-volume transfer.}
\textbf{(a)} The same saved Qwen3-VL SFT points decoded by frozen
microSAM and SAM 3.
\textbf{(b)} AP$_{50}$ under the standard instruction and five held-out
mitochondria phrasings for the existing SFT and SFT+GRPO checkpoints.
\textbf{(c)} Standard and morphology-only instructions for unadapted
Qwen3-VL and the existing SFT checkpoint; dashed lines show the mean over
the three morphology-only phrasings.
\textbf{(d)} Point-in-mask precision, object recall, and AP$_{50}$ for
the unchanged combined three-epoch Qwen3-VL SFT adapter on ``All data''
and the independent larval-fly volume.}
\label{fig:robustness}
\end{figure*}

The combined adapter also transfers beyond the three public datasets, but
performance declines on the independent larval-fly volume
(Fig.~\ref{fig:robustness}d). Point-in-mask precision decreases
moderately, while object recall and downstream segmentation fall more
substantially; AP$_{50}$ drops from $0.712$ on ``All data'' to $0.464$
on the independent volume. The adapter therefore transfers to the independent EM volume, but with
lower localization and segmentation performance than on the public test
data. Full results are reported in Supplementary
Section~\ref{app:larvalfly}.

\subsection{Coordinates remain usable across decoders and instruction phrasings}
\label{subsec:robustness}

The saved Qwen3-VL SFT coordinates remain usable with a second
segmentation backend. Decoding the same points with SAM 3 gives
AP$_{50}$ $0.698$, close to the $0.712$ obtained with microSAM
(Fig.~\ref{fig:robustness}a). The coordinate hand-off is therefore not
specific to the microSAM decoder, although microSAM remains slightly
better in this setting.

The adapted Qwen3-VL models are also stable to unseen formulations
of the mitochondria request. Across five phrasings excluded from training,
AP$_{50}$ ranges from $0.712$ to $0.726$ after SFT and from $0.729$ to
$0.735$ after SFT+GRPO (Fig.~\ref{fig:robustness}b). These variations
are small relative to the adaptation gain itself, indicating that the
result is not tied to the exact wording of the standard instruction.

Removing the word ``mitochondria'' gives a harder test. With three
morphology-only descriptions, the unadapted model remains close to its
standard-instruction result on average ($0.248$ versus $0.247$
AP$_{50}$), while the existing SFT checkpoint averages $0.668$ rather
than $0.712$ (Fig.~\ref{fig:robustness}c). Performance is also more variable across these descriptions (Supplementary Section~\ref{app:prompt}), but the
adapted model retains much of its downstream segmentation performance when
the target is described through morphology alone.

\subsection{Attributing the gain: supervision, data, adapter capacity, and backend}
\label{subsec:ablations}

\begin{figure*}[t]
\centering
\includegraphics[width=\textwidth]{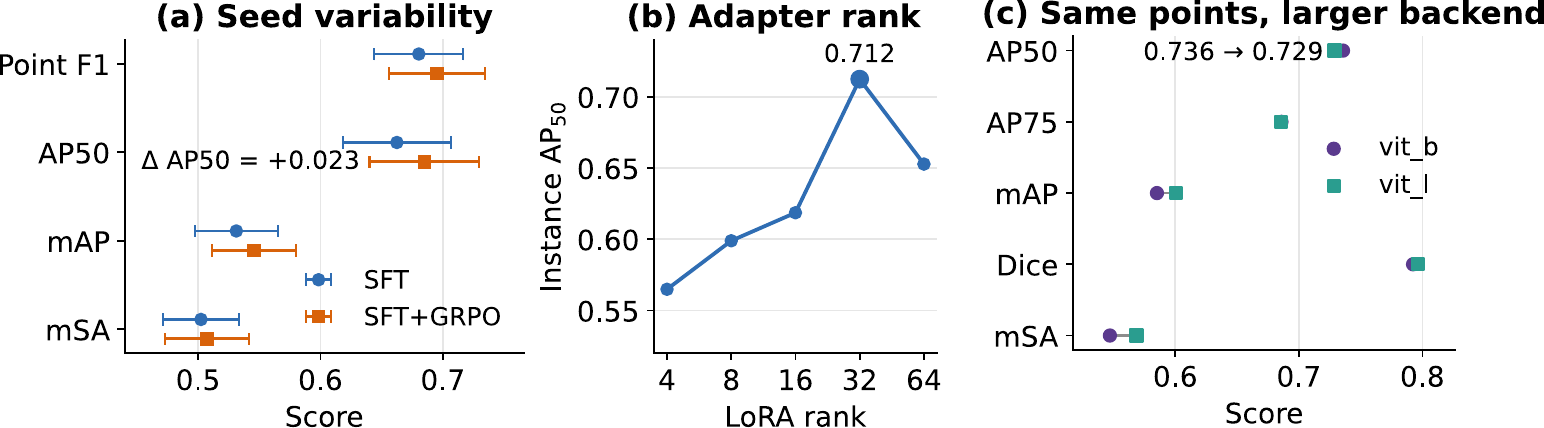}
\caption{\textbf{Adaptation and backend ablations.}
\textbf{(a)} Mean $\pm$ standard deviation across seeds $7$, $42$, and
$123$ for Qwen3-VL SFT and SFT+GRPO.
\textbf{(b)} AP$_{50}$ across tested LoRA ranks.
\textbf{(c)} The same Qwen3-VL GRPO points decoded with the
\texttt{vit\_b} and \texttt{vit\_l} microSAM backends.}
\label{fig:ablations}
\end{figure*}

\noindent
\textbf{Supervised fine-tuning provides most of the gain; GRPO adds a
smaller refinement.}
Table~\ref{tab:overall} retains the primary seed-42 Qwen3-VL run
because the other trained models in the table are also represented by
single runs; Fig.~\ref{fig:ablations}a separately reports the
three-seed Qwen3-VL mean and variability.
Across three seeds, mean AP$_{50}$ increases from $0.662\pm0.044$ after
SFT to $0.685\pm0.045$ after GRPO (Fig.~\ref{fig:ablations}a). Point F1,
mAP, and mSA also increase, but the AP$_{50}$ gain of $0.023$ is modest
relative to the between-seed variation.  Note that the GRPO reward combines
five terms whose contributions were not ablated individually, so we cannot
attribute this modest gain to a particular reward component.

\noindent
\textbf{Performance scales with mask-derived supervision.}
On the fixed Lucchi++ test set, SFT AP$_{50}$ rises from $0.297$ with $5$
source slices ($105$ crops) to $0.717$ with all $92$ slices ($2{,}869$
crops; Fig.~\ref{fig:panel}a).

\noindent
\textbf{Adapter capacity is non-monotonic.}
AP$_{50}$ increases with LoRA rank to $0.712$ at rank $32$, then falls
to $0.653$ at rank $64$ (Fig.~\ref{fig:ablations}b).

\noindent
\textbf{A larger microSAM backend does not improve AP$_{50}$.}
Decoding the same GRPO points with \texttt{vit\_l\_em\_organelles}
instead of \texttt{vit\_b\_em\_organelles} changes AP$_{50}$ from
$0.736$ to $0.729$, while AP$_{75}$, mAP, Dice, and mSA increase
slightly (Fig.~\ref{fig:ablations}c).

\section{Conclusion}
\label{sec:conclusion}

We investigated whether open-weight MLLMs can be adapted to generate point
prompts from natural-language requests for electron microscopy segmentation
while leaving the downstream segmentation model frozen. Our experiments show
that this is feasible. Zero-shot localization is weak, but SFT on
mask-derived centroid targets substantially improves point generation, while
GRPO provides a smaller refinement. Importantly, the improvement is not
specific to Qwen3-VL. Although the primary Qwen3-VL result remains below APG,
the matched one-epoch comparison shows that InternVL3.5 reaches similar
performance and Molmo2 exceeds it, indicating that the effectiveness of the
adaptation also depends on the underlying MLLM.  The learned localization also transfers to excluded public datasets and
to a separate larval-fly volume, although performance declines on the
independent specimen. At the same time, none of the
adapted MLLMs surpasses the centroid-heatmap detector in AP$_{50}$ for this
mitochondria task. Our point-level and proposal-suppression analyses show that
the detector's higher AP$_{50}$ is driven in part by producing many additional
point proposals, which increases object coverage rather than agreement with
the annotated centroid set.

More broadly, this work shows that MLLMs can be adapted to serve as point
generators for an EM target while retaining natural
language as the interface. Their explicit image-space coordinates separate
localization from mask decoding at inference, allowing the predicted points
to be inspected directly and passed to point-prompted segmenters such as
microSAM. This provides a route from a natural-language request to an
executable spatial prompt without modifying the downstream segmentation
model, and makes the localization stage independently accessible rather than
embedding it within the segmenter itself.

Future work should extend the supervision beyond mitochondria to several
organelles and cellular structures, requiring one adapted model to respond
differently to requests for different targets rather than to alternative
phrasings of a single target. It should also explore learning directly in three-dimensional
space from volumetric annotations, rather than reducing volumes to
independently processed two-dimensional slices. A further direction is
integration into closed-loop microscopy systems in which predicted
locations guide subsequent acquisition or
analysis~\cite{kalinin2023automatedexperimentation}.



\section*{Acknowledgements}
This project was funded by a Wellcome Trust Investigator Award (Ref: 205038/Z/16/Z), an ERC grant (Ref: ERC-2018-COG: 819650), a BBSRC Grant (Ref: APP26929, ``Population connectomics'') and MRC LMB core funding.

\bibliographystyle{splncs04}
\bibliography{references_seg, references_mendeley}

%

\clearpage
\section*{Supplementary Material}
\section{Implementation and training details}
\label{app:implementation}

All experiments use $256\times256$ crops. Images are opened as RGB and passed
through each model's default processor, so model-specific resizing and
normalization remain unchanged. The evaluated checkpoints are
\texttt{Qwen/\allowbreak Qwen3-VL-8B-Instruct},
\texttt{OpenGVLab/InternVL3\_5-8B-HF}, and
\texttt{allenai/Molmo2-8B}. Qwen3-VL and InternVL3.5 are loaded in
bfloat16. Molmo2 uses its default automatic dtype selection, with bfloat16
as the fallback. No quantization is used.

Inference is deterministic. We use greedy decoding with sampling disabled
and a maximum of $192$ generated tokens. Standard evaluation uses the
instruction ``Identify all mitochondria in this electron microscopy image.''
Training cycles through $35$ mitochondria instructions. The five instructions
used for the wording-robustness experiment are excluded from this training
pool.

XML-style point tokens are parsed from the generated text, and Molmo2's
native point-token format is converted to the same normalized representation.
Coordinates are rounded to four decimal places. Malformed or out-of-range
tokens are skipped rather than clipped. If no valid point token is found, the
prediction is treated as an empty point set. Duplicate coordinates are not
removed before point evaluation. They therefore count as repeated predictions,
although masks produced by repeated prompts can later be suppressed by the
mask-overlap rule. The parsed point count, rather than the model's textual
integer, determines point-count MAE and the GRPO count reward.

Table~\ref{tab:supp_mllm_hparams} summarizes the training settings used for
supervised fine-tuning and reward optimization.

\FloatBarrier
\begin{table}[ht!]
\centering\small
\caption{Core training settings used for the MLLM experiments. The primary
Qwen3-VL result uses three SFT epochs; the matched cross-backbone comparison
uses one epoch for every model.}
\label{tab:supp_mllm_hparams}
\begin{tabular}{lcc}
\toprule
Setting & SFT & GRPO \\
\midrule
LoRA rank / $\alpha$ / dropout & $32/64/0.05$ & inherited from SFT \\
Learning rate & $2\times10^{-5}$ & $5\times10^{-7}$ \\
Per-device batch size & $1$ & $1$ \\
Gradient accumulation & $8$ & $4$ \\
Epochs & $3$ primary; $1$ cross-model & $1$ \\
Schedule / warmup & cosine / $0.05$ & implementation default \\
Generated responses per crop & -- & $4$ \\
Maximum completion length & $192$ & $192$ \\
Temperature / top-$p$ / $\beta$ & -- & $1.0/0.95/0.04$ \\
\bottomrule
\end{tabular}
\end{table}
\FloatBarrier

For Qwen3-VL and InternVL3.5, LoRA targets the available attention and
feed-forward projection layers. Molmo2 uses the corresponding available
attention and feed-forward linear modules exposed by its implementation.
The main Qwen3-VL rank-32 adapter contains $87{,}293{,}952$ trainable
parameters out of $8{,}854{,}417{,}648$ total parameters
($0.986\%$). The Molmo2 rank-32 adapter contains $83{,}252{,}224$
trainable parameters out of $8{,}744{,}955{,}344$ total parameters
($0.952\%$).

\section{Dataset split sizes}
\label{app:data_stats}

Table~\ref{tab:supp_dataset_sizes} reports the crop counts used by the
combined adaptation and the fixed test sets. Splits are defined at the
source-slice level before tiling into crops. The combined validation set contains $752$ crops.

\FloatBarrier
\begin{table}[ht!]
\centering\small
\caption{Training and test crop counts for the three public datasets.}
\label{tab:supp_dataset_sizes}
\begin{tabular}{lrr}
\toprule
Dataset & Training crops & Test crops \\
\midrule
Lucchi++ & 2,869 & 342 \\
VNC & 433 & 79 \\
MitoEM-H & 2,655 & 332 \\
\midrule
Combined & 5,957 & 753 \\
\bottomrule
\end{tabular}
\end{table}
\FloatBarrier

\section{Reward and target-format details}
\label{app:reward}

The GRPO reward combines five terms,
\begin{equation}
r = 0.10\,r_{\text{fmt}} + 0.15\,r_{\text{cnt}} + 0.20\,r_{\text{pt}} + 0.20\,r_{\text{fg}}
  + 0.35\,r_{\text{IoU}},
\end{equation}
where $r_{\text{fmt}}$ rewards a parseable point response, or the correct empty response when no target is
present, and
\[
r_{\text{cnt}}=\exp\!\left[-\frac{1}{2}
\left(\frac{\hat N_{\mathrm{pt}}-N}{1.5}\right)^2\right]
\]
penalizes parsed point-count error relative to the ground-truth count $N$. For nonempty point sets,
$r_{\text{pt}}$ greedily matches predicted points to unused ground-truth centroids by normalized Euclidean
distance and computes
\[
r_{\text{pt}}=
\frac{1}{\max(\hat N_{\mathrm{pt}},N)}
\sum_{(p,g)\in\mathcal{M}_{\mathrm{pt}}}
\exp\!\left[-\frac{1}{2}
\left(\frac{\lVert p-g\rVert_2}{0.10}\right)^2\right].
\]
The score is $1$ when both point sets are empty and $0$ when exactly one is empty. Dividing by the larger set
size means that unmatched predictions and targets lower the reward. When predicted points are present, the
foreground term $r_{\text{fg}}$ is the fraction that fall inside labeled foreground. When no points are
predicted, this term is $1$ only for an empty ground-truth foreground and $0$ otherwise. The mask term uses
Hungarian assignment and, for nonempty mask sets, is
\[
r_{\text{IoU}}=
\frac{1}{\max(\hat N_{\mathrm{mask}},N_{\mathrm{mask}})}
\sum_{(\hat m,m)\in\mathcal{M}_{\mathrm{mask}}}
\operatorname{IoU}(\hat m,m).
\]
It uses the same empty-set convention as the point term. The
reported task requests positive point tokens for mitochondria, so the five terms above define the intended
reward. The weights were selected using validation data and then fixed before test evaluation.

A multiplicative degenerate-output penalty then suppresses malformed or repeated responses, taking value $0$
for leaked placeholders or responses composed only of the fixed template coordinates, $0.25$ when any rounded
coordinate is repeated, and $1$ otherwise.

The target text states the mitochondria count in words followed by one XML-style point token per instance.
Each token stores normalized \texttt{x} and \texttt{y} attributes and an instance label in the \texttt{alt}
attribute. Coordinates lie in $[0,1]$ in $(x,y)$ order, with the origin at the top left, $x$ increasing
rightward, and $y$ increasing downward. For cross-entropy SFT, the target tokens are serialized
deterministically: instance-labeled datasets use ascending numeric instance identifiers, and semantic-only
datasets use connected-component scan order. The objective therefore operates on a deterministic
serialization of the target points rather than a permutation-invariant set loss.

\section{Point and centroid diagnostics}
\label{app:point_mask}
Table~\ref{tab:supp_point_mask} reports crop-aligned point diagnostics over the $753$ ``All data'' test crops.
Diagnostic point F1 uses the normalized threshold $0.10$ and is distinct from the primary point F1 at
$0.05$. Point-count MAE is computed from valid parsed point tokens before mask decoding. Arithmetic centroids
lie inside their source masks for $1{,}747$ of $1{,}768$ instances, showing that invalid centroid targets are uncommon in the evaluated
instances.

\FloatBarrier
\begin{table}[ht!]
\centering\small\setlength{\tabcolsep}{4pt}
\caption{Crop-aligned point diagnostics over the $753$ ``All data'' test crops. Diagnostic point F1 uses
threshold $0.10$; point-count MAE uses valid parsed point tokens.}
\label{tab:supp_point_mask}
\resizebox{\textwidth}{!}{%
\begin{tabular}{lcccccc}
\toprule
Model and stage & Diag.\ F1 @0.10 & Point-in-mask prec. & Object recall & Point-count MAE & Centroids in masks & Instances \\
\midrule
Qwen3-VL zero-shot & 0.170 & 0.293 & 0.201 & 1.182 & 0.988 & 1747/1768 \\
Qwen3-VL SFT       & 0.747 & 0.852 & 0.718 & 0.603 & 0.988 & 1747/1768 \\
Qwen3-VL SFT+GRPO  & 0.765 & 0.854 & 0.742 & 0.566 & 0.988 & 1747/1768 \\
InternVL3.5 SFT    & 0.806 & 0.897 & 0.791 & 0.534 & 0.988 & 1747/1768 \\
Molmo2 SFT         & 0.846 & 0.918 & 0.840 & 0.465 & 0.988 & 1747/1768 \\
\bottomrule
\end{tabular}}
\end{table}
\FloatBarrier

\section{Per-condition breakdown of the point generators}
\label{app:percondition}
Table~\ref{tab:supp_qwen_condition} reports the combined Qwen3-VL model
evaluated separately on the three source datasets. MitoEM-H has the lowest score and also the most mitochondria per crop and
the smallest median annotated instance area in pixels; these descriptive
differences are consistent with the result but do not identify a single cause. Since acquisition resolutions differ across datasets, the pixel-area values
should not be interpreted as physical mitochondrial sizes.

\FloatBarrier
\begin{table}[ht!]
\centering\small
\caption{The combined Qwen3-VL model evaluated by source condition. Macro is the unweighted mean of the three
conditions; ``All data'' is the average over all $753$ test crops. Values are AP$_{50}$.}
\label{tab:supp_qwen_condition}
\begin{tabular}{lc|c|c|c|c}
\toprule
Stage & Lucchi++ & MitoEM-H & VNC & Macro & All data \\
\midrule
SFT      & 0.831 & 0.605 & 0.647 & 0.695 & 0.712 \\
SFT+GRPO & 0.843 & 0.630 & 0.714 & 0.729 & 0.736 \\
\bottomrule
\end{tabular}
\end{table}
\FloatBarrier

The ground-truth test-set statistics in Table~\ref{tab:supp_dataset_morphology} provide context
for this descriptive comparison. MitoEM-H has the largest mean and median number of mitochondria per crop
and the smallest median instance area. These statistics describe the evaluated labels and do not establish
that density or object size causes the lower segmentation score.

\FloatBarrier
\begin{table}[ht!]
\centering\small
\caption{Ground-truth instance density and size on the public test crops. Instance area is measured in pixels
within the $256\times256$ crops.}
\label{tab:supp_dataset_morphology}
\begin{tabular}{l|r|r|r|r|r}
\toprule
Dataset & Crops & Instances & Mean inst./crop & Median inst./crop & Median area (px) \\
\midrule
Lucchi++ & 342 & 715 & 2.091 & 2 & 2,109 \\
MitoEM-H & 332 & 899 & 2.708 & 3 & 1,108 \\
VNC & 79 & 154 & 1.949 & 2 & 1,390 \\
\bottomrule
\end{tabular}
\end{table}
\FloatBarrier

\section{Adapter-capacity ablation}
\label{app:lora}
Table~\ref{tab:supp_lora_rank} reports the complete adapter-rank ablation.
Performance improves through rank $32$ but falls at rank $64$, so adapter capacity helps non-monotonically
and rank $32$ is the strongest tested setting rather than simply the largest adapter.

\FloatBarrier
\begin{table}[ht!]
\centering\small
\caption{Combined Qwen3-VL SFT sensitivity to LoRA rank, same training data and $753$ test crops through
frozen \texttt{vit\_b\_em\_organelles}. Values are AP$_{50}$.}
\label{tab:supp_lora_rank}
\begin{tabular}{c|c}
\toprule
LoRA rank & AP$_{50}$ \\
\midrule
4  & 0.565 \\
8  & 0.599 \\
16 & 0.619 \\
32 & 0.712 \\
64 & 0.653 \\
\bottomrule
\end{tabular}
\end{table}
\FloatBarrier

\section{Backend-portability control}
\label{app:sam3}
Table~\ref{tab:supp_sam3} executes the same saved Qwen3-VL SFT points through two frozen decoders. Direct
SAM 3 text prompting does not recover usable mitochondria masks in this EM configuration, whereas passing the
saved points to SAM 3 reaches AP$_{50}$ $0.698$, close to the $0.712$ obtained with microSAM. This supports the
portability of the explicit coordinate hand-off while also supporting the microscopy-adapted backend for the
principal comparison. We do not generalize the near-zero direct SAM 3 result beyond this prompt and setting.

\FloatBarrier
\begin{table}[ht!]
\centering\small\setlength{\tabcolsep}{4pt}
\caption{Backend-portability control on the same $753$ ``All data'' test crops. The two Qwen3-VL rows use identical
saved SFT point predictions and differ only in the frozen downstream decoder. Direct SAM 3 prompting uses the
text prompt \texttt{mitochondrion}.}
\label{tab:supp_sam3}
\resizebox{\textwidth}{!}{%
\begin{tabular}{llcccccc}
\toprule
Input source & Backend & Dice & AP$_{50}$ & AP$_{75}$ & mAP & mSA & Count MAE \\
\midrule
SAM 3 text prompt    & SAM 3     & 0.008 & 0.008 & 0.008 & 0.008 & 0.008 & 2.348 \\
Qwen3-VL SFT points & SAM 3     & 0.761 & 0.698 & 0.642 & 0.564 & 0.536 & 0.607 \\
Qwen3-VL SFT points & microSAM & 0.776 & 0.712 & 0.659 & 0.570 & 0.538 & 0.606 \\
\bottomrule
\end{tabular}}
\end{table}
\FloatBarrier

\section{Held-out Lucchi++ training extension}
\label{app:transfer_extension}

The leave-one-dataset-out comparison in the main paper uses one SFT epoch
for every held-out condition. For Lucchi++, we additionally continued the
same held-out adapter to three SFT epochs. AP$_{50}$ increased from
$0.694$ after one epoch to $0.712$ after three epochs. This extension was
not run for the VNC or MitoEM-H holdouts, so it is reported only as a
training-duration check rather than as a matched transfer comparison.

\section{Independent-volume evaluation}
\label{app:larvalfly}
Table~\ref{tab:supp_larvalfly} evaluates the unchanged combined three-epoch
Qwen3-VL SFT adapter on a separate in-house larval-fly EM volume absent
from adaptation. The volume is a \textit{Drosophila} larval sample imaged
by FIB-SEM at $8\times8\times8$~nm isotropic resolution. Evaluation uses
a fixed one-percent subset of the generated crops ($131$ crops), selected
before model evaluation. The adapter is used without further training and
its points are decoded by frozen
\texttt{vit\_b\_em\_organelles}. Diagnostic point F1 uses threshold $0.10$.

\FloatBarrier
\begin{table}[ht!]
\centering\small
\caption{Independent-volume evaluation on the fixed one-percent in-house larval-fly subset.}
\label{tab:supp_larvalfly}
\resizebox{\textwidth}{!}{%
\begin{tabular}{lc|c|c|c|c|c|c|c|c|c}
\toprule
Subset & Crops & Diag.\ F1 & Point-in-mask & Object recall & Dice & AP$_{50}$ & AP$_{75}$ & mAP & mSA & Count MAE \\
\midrule
Larval-fly 1\% & 131 & 0.545 & 0.753 & 0.462 & 0.567 & 0.464 & 0.438 & 0.379 & 0.372 & 1.214 \\
\bottomrule
\end{tabular}}
\end{table}
\FloatBarrier

\section{Larger microSAM backend control}
\label{app:large_backend}

Table~\ref{tab:supp_large_backend} compares the two microSAM backends using
the same saved Qwen3-VL SFT+GRPO point predictions. We execute these points with the
frozen \texttt{vit\_b\_em\_organelles} and
\texttt{vit\_l\_em\_organelles} checkpoints. The larger backend lowers the
primary AP$_{50}$ from $0.736$ to $0.729$, while AP$_{75}$, mAP, Dice, and
mSA increase slightly. Count MAE is unchanged. The result therefore does
not support a primary-threshold gain from increasing the segmentation
backbone size.

\FloatBarrier
\begin{table}[ht!]
\centering\small
\caption{Backend-capacity control using the same saved Qwen3-VL SFT+GRPO
point predictions on all $753$ test crops.}
\label{tab:supp_large_backend}
\begin{tabular}{l|c|c|c|c|c|c}
\toprule
Backend & Dice & AP$_{50}$ & AP$_{75}$ & mAP & mSA & Count MAE \\
\midrule
\texttt{vit\_b\_em\_organelles} & 0.792 & 0.736 & 0.678 & 0.585 & 0.547 & 0.564 \\
\texttt{vit\_l\_em\_organelles} & 0.797 & 0.729 & 0.686 & 0.601 & 0.569 & 0.564 \\
\bottomrule
\end{tabular}
\end{table}
\FloatBarrier

\section{Prompt-phrasing sensitivity}
\label{app:prompt}

The wording-robustness experiment uses five mitochondria instructions that
do not appear verbatim in the $35$-instruction training pool. These
same-target phrasings test sensitivity to instruction wording rather than
to a different biological target.

\begin{enumerate}
\item[\texttt{pv01}] Please identify every mitochondrion in this EM crop and
return one normalized point inside each object.
\item[\texttt{pv02}] Inspect the micrograph and mark all mitochondrial
profiles with one center point per instance.
\item[\texttt{pv03}] For this electron microscopy patch, give point
annotations for each visible mitochondrion; include none if absent.
\item[\texttt{pv04}] Find every complete or partial mitochondrial structure
in the crop and provide one point token per structure.
\item[\texttt{pv05}] Act as an EM annotator: count the mitochondria and place
a normalized point near the middle of each one.
\end{enumerate}

Table~\ref{tab:supp_prompt_detailed} reports the corresponding mask results.
SFT is not reduced on average by the held-out formulations. GRPO decreases
slightly, from AP$_{50}$ $0.736$ under the standard instruction to a five-prompt
mean of $0.732$. Across the five requests, AP$_{50}$ ranges from $0.712$ to
$0.726$ after SFT and from $0.729$ to $0.735$ after SFT+GRPO.

\FloatBarrier
\begin{table}[ht!]
\centering\small\setlength{\tabcolsep}{3.5pt}
\caption{Sensitivity to five held-out formulations of the mitochondria
phrasings. All rows use the same $753$ test crops and frozen
\texttt{vit\_b\_em\_organelles} backend.}
\label{tab:supp_prompt_detailed}
\resizebox{\textwidth}{!}{%
\begin{tabular}{llc|c|c|c}
\toprule
Stage & Phrasing & Dice $\uparrow$ & AP$_{50}$ $\uparrow$  & mAP $\uparrow$ & Count MAE $\downarrow$ \\
\midrule
SFT & Standard & 0.776 & 0.712 & 0.570 & 0.606 \\
\cmidrule(lr){2-6} 

SFT & \texttt{pv01} & 0.777 & 0.715 & 0.571 & 0.603 \\
SFT & \texttt{pv02} & 0.782 & 0.726 & 0.580 & 0.579 \\
SFT & \texttt{pv03} & 0.775 & 0.712 & 0.570 & 0.614 \\
SFT & \texttt{pv04} & 0.785 & 0.723 & 0.576 & 0.575 \\
SFT & \texttt{pv05} & 0.778 & 0.717 & 0.571 & 0.620 \\
\cmidrule(lr){3-6} 
SFT & Unseen mean & 0.779 & 0.719 & 0.574 & 0.598 \\
\midrule
SFT+GRPO & Standard & 0.792 & 0.736 & 0.585 & 0.564 \\
\cmidrule(lr){2-6} 

SFT+GRPO & \texttt{pv01} & 0.788 & 0.730 & 0.580 & 0.575 \\
SFT+GRPO & \texttt{pv02} & 0.788 & 0.734 & 0.584 & 0.563 \\
SFT+GRPO & \texttt{pv03} & 0.787 & 0.729 & 0.580 & 0.571 \\
SFT+GRPO & \texttt{pv04} & 0.789 & 0.735 & 0.582 & 0.562 \\
SFT+GRPO & \texttt{pv05} & 0.789 & 0.732 & 0.581 & 0.570 \\
\cmidrule(lr){3-6} 
SFT+GRPO & Unseen mean & 0.788 & 0.732 & 0.581 & 0.568 \\
\bottomrule
\end{tabular}}
\end{table}
\FloatBarrier

\paragraph{Morphology-only phrasings.}
The five phrasings above explicitly name mitochondria. We therefore test
whether the performance depends on naming the target, using three `morphology-only' phrasings that describe it only through visual morphology. These phrasings
are used only at inference with the unadapted Qwen3-VL model and the
existing SFT checkpoint, without further training. Evaluation uses the same $753$ test crops. The points are decoded with the frozen
\texttt{vit\_b\_em\_organelles} backend:

\begin{description}
\item[\texttt{morph1}]
Locate every elongated, oval, or bean-shaped enclosed structure with
visible internal folds or parallel ridges. Return one normalized point
near the center of each structure.

\item[\texttt{morph2}]
Mark each bounded object with a darker outer rim and repeated folded or
stripe-like internal texture. Include partial objects at the image border
and return one point per object.

\item[\texttt{morph3}]
Find all compact or elongated enclosed regions containing several internal
curved membranes or ridges. Return one normalized point near the middle of
each region.
\end{description}

\begin{table}[t]
\centering
\small
\caption{\textbf{Morphology-only phrasing control on ``All data'' at inference.}
Point F1 uses $\tau=0.05$.}
\label{tab:supp_morphology}
\begin{tabular}{lcccc}
\toprule
& \multicolumn{2}{c}{Unadapted}
& \multicolumn{2}{c}{SFT checkpoint} \\
Phrasing & Point F1 & AP$_{50}$ & Point F1 & AP$_{50}$ \\
\midrule
Morph-1 & 0.036 & 0.180 & 0.664 & 0.606 \\
Morph-2 & 0.053 & 0.287 & 0.713 & 0.730 \\
Morph-3 & 0.042 & 0.276 & 0.707 & 0.668 \\
\midrule
Mean    & 0.044 & 0.248 & 0.695 & 0.668 \\
\bottomrule
\end{tabular}
\end{table}

Table~\ref{tab:supp_morphology} reports the corresponding point and
segmentation results. After SFT, the morphology-only phrasings are more variable than the five
held-out mitochondria phrasings (AP$_{50}$ $0.606$--$0.730$ versus
$0.712$--$0.726$). The unadapted mean is essentially unchanged from the
standard instruction ($0.248$ versus $0.247$), whereas the existing SFT
checkpoint reaches $0.668$ on the same morphology-only inputs. This
control changes only the inference phrasing and does not test
discrimination among biological targets.

\section{Single-dataset point generators}
\label{app:singledataset}
Table~\ref{tab:supp_single_dataset} reports the Qwen3-VL adapters trained
separately on Lucchi++, VNC, and MitoEM-H using seed $42$ and three SFT
epochs. The SFT+GRPO results continue from the corresponding SFT
checkpoints with one GRPO epoch. The VNC-only adapter is the weakest of the three. Its smaller training set may contribute to this, although
the experiment does not separate training-set size from differences in morphology or imaging condition.

\FloatBarrier
\begin{table}[ht!]
\centering\small
\caption{Qwen3-VL adapters trained and tested separately on each dataset
using seed $42$ and three SFT epochs. SFT+GRPO adds one GRPO epoch. Values are AP$_{50}$.}
\label{tab:supp_single_dataset}
\begin{tabular}{lcc}
\toprule
Dataset & SFT & SFT+GRPO \\
\midrule
Lucchi++ & 0.709 & 0.711 \\
MitoEM-H & 0.531 & 0.567 \\
VNC      & 0.436 & 0.463 \\
\bottomrule
\end{tabular}
\end{table}
\FloatBarrier

The all-slices point in the data-scaling experiment is produced by a
separately trained, fixed-test scaling checkpoint and reaches AP$_{50}$
$0.717$. Table~\ref{tab:supp_single_dataset} reports the primary
single-dataset checkpoint at $0.709$. The two values therefore come from
distinct training runs evaluated on the same fixed Lucchi++ test set.

\section{Centroid-heatmap detector}
\label{app:heatmap}

The centroid-heatmap baseline is a $488{,}417$-parameter U-Net-style
detector trained on the same crops and centroid targets used for MLLM
adaptation. Each target point is rendered as a Gaussian with standard
deviation $3$ pixels. The combined detector uses $5{,}957$ training crops,
$752$ validation crops, and $753$ test crops, with seed $42$, batch size
$32$, and at most $12$ epochs. The positive-class weight is $20$.

At inference, local maxima above $0.25$ are extracted from the predicted
heatmap. Peaks must be separated by at least $8$ pixels, and at most $64$
points are retained per crop. The resulting coordinates are passed to the
same frozen \texttt{vit\_b\_em\_organelles} backend and evaluated with the
same crop-level segmentation metrics as the MLLM points. The point F1
values reported in the main paper use the common normalized matching
threshold $\tau=0.05$. The leave-one-dataset-out heatmap detectors use the
same held-out split definitions as the corresponding Qwen3-VL adapters.

Table~\ref{tab:supp_heatmap} reports the combined and leave-one-dataset-out
point and segmentation diagnostics.

\FloatBarrier
\begin{table}[ht!]
\centering\small
\caption{Additional centroid-heatmap diagnostics. Point F1 uses
$\tau=0.05$.}
\label{tab:supp_heatmap}
\begin{tabular}{lc|c|c|c|c|c}
\toprule
Setting & Point F1 & PiM & Obj.\ recall & AP$_{50}$ & mSA & Count MAE \\
\midrule
Combined & 0.674 & 0.655 & 0.964 & 0.904 & 0.456 & 1.483 \\
Held-out Lucchi++ & 0.424 & 0.383 & 0.920 & 0.888 & 0.282 & 3.570 \\
Held-out VNC & 0.385 & 0.325 & 0.884 & 0.804 & 0.194 & 3.722 \\
Held-out MitoEM-H & 0.549 & 0.660 & 0.749 & 0.708 & 0.387 & 1.108 \\
\bottomrule
\end{tabular}
\end{table}
\FloatBarrier

\section{Supervised U-Net reference}
\label{app:unet}
Table~\ref{tab:supp_unet_condition} reports the single combined U-Net evaluated by source condition; these are
not dataset-specific U-Nets but the one combined model evaluated separately by dataset. Instances are connected components of
the thresholded foreground map, ranked by mean foreground probability.

\FloatBarrier
\begin{table}[ht!]
\centering\small
\caption{The single combined U-Net evaluated by source condition. ``All data'' is the average over all $753$ test
crops.}
\label{tab:supp_unet_condition}
\begin{tabular}{lccccccc}
\toprule
Condition & Crops & Dice & AP$_{50}$ & AP$_{75}$ & mAP & mSA & Count MAE \\
\midrule
Lucchi++ & 342 & 0.949 & 0.944 & 0.934 & 0.890 & 0.837 & 0.231 \\
MitoEM-H & 332 & 0.865 & 0.834 & 0.741 & 0.682 & 0.560 & 0.756 \\
VNC      & 79  & 0.824 & 0.863 & 0.693 & 0.637 & 0.490 & 0.835 \\
\midrule
All data & 753 & 0.899 & 0.887 & 0.824 & 0.772 & 0.679 & 0.526 \\
\bottomrule
\end{tabular}
\end{table}
\FloatBarrier

The combined reference is a five-level two-dimensional U-Net with $32$ base channels and $7{,}851{,}969$
parameters. Each block uses convolution, GroupNorm, and SiLU activations, with max-pooling in the encoder and
bilinear upsampling with skip concatenation in the decoder. Training uses AdamW with learning rate $10^{-3}$, weight decay $10^{-4}$,
batch size $8$, and at most $50$ epochs. Early stopping uses patience $10$ and minimum
validation-loss improvement $10^{-4}$. The objective is binary cross-entropy with logits
plus Dice loss, and augmentation consists of random horizontal and vertical flips.
The checkpoint with the lowest validation loss is retained. Foreground probability is
thresholded at $0.5$ and converted to instances by two-dimensional connected components,
without watershed or minimum-size filtering.

\section{Point and downstream mask diagnostics}
\label{app:point_ap}

Point F1 and downstream AP$_{50}$ evaluate different stages of the
point-prompting pipeline. Point F1 compares the predicted point set with
the annotated centroids using one-to-one matching at $\tau=0.05$, so
unmatched predictions and missed targets reduce the score. AP$_{50}$ is
computed after the points have been decoded into masks; for
point-prompted methods, these masks are scored by microSAM's predicted-IoU
confidence before the precision--recall calculation. The two metrics can
therefore respond differently when additional point proposals increase
object coverage.

\paragraph{Combined heatmap detector.}
We first vary the peak-NMS minimum distance for the combined
centroid-heatmap detector evaluated on ``All data,'' without retraining
the detector. The saved heatmap predictions are decoded using minimum
distances $\{8,12,16,24,32,48\}$ pixels, with all other post-processing
and evaluation settings unchanged. Radius $8$ is the reported operating
point and exactly reproduces the main ``All data'' result: point F1
$0.674$, AP$_{50}$ $0.904$, and instance-count MAE $1.483$. Point F1
uses the primary matching threshold $\tau=0.05$ throughout.

Increasing the suppression radius removes nearby proposals and improves
agreement with the annotated point set and instance count while leaving
AP$_{50}$ comparatively stable. At radius $24$, point F1 increases to
$0.681$ and count MAE decreases to $1.378$, while AP$_{50}$ remains
$0.903$. At radius $48$, point F1 reaches $0.690$ and count MAE
$1.094$, while AP$_{50}$ decreases to $0.885$.
Table~\ref{tab:supp_combined_nms} reports the complete sweep.

\begin{table}[ht!]
\centering\small
\caption{\textbf{Peak-NMS sensitivity of the combined centroid-heatmap
detector on ``All data.''} The detector is not retrained. Radius $8$ is
the setting used for the reported result. Point F1 uses $\tau=0.05$.}
\label{tab:supp_combined_nms}
\begin{tabular}{cccc}
\toprule
Radius & Point F1 & AP$_{50}$ & Count MAE \\
\midrule
8  & 0.674 & 0.904 & 1.483 \\
12 & 0.677 & 0.904 & 1.466 \\
16 & 0.679 & 0.904 & 1.434 \\
24 & 0.681 & 0.903 & 1.378 \\
32 & 0.683 & 0.900 & 1.303 \\
48 & 0.690 & 0.885 & 1.094 \\
\bottomrule
\end{tabular}
\end{table}

\paragraph{Leave-one-dataset-out control.}

\begin{figure*}[ht!]
\centering
\includegraphics[width=0.32\textwidth]{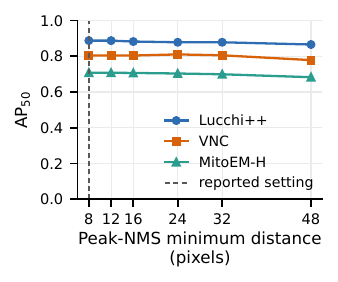}
\hfill
\includegraphics[width=0.32\textwidth]{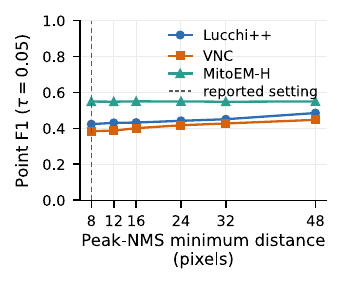}
\hfill
\includegraphics[width=0.32\textwidth]{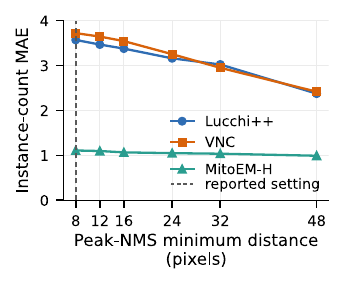}
\caption{\textbf{Sensitivity of the leave-one-dataset-out
centroid-heatmap detectors to peak suppression.} The minimum peak
distance is varied for the three detectors; $8$ pixels is the setting
used for the reported transfer results. Point F1 uses the primary
matching threshold $\tau=0.05$. Increasing suppression generally
increases Point F1 and reduces instance-count MAE, while downstream
AP$_{50}$ remains comparatively stable across the sweep.}
\label{fig:supp_heatmap_nms}
\end{figure*}

We observe the same overall pattern for the three
leave-one-dataset-out heatmap detectors used in the transfer comparison.
For each detector, we vary only the peak-NMS minimum distance while
keeping the saved heatmap predictions, frozen microSAM backend, and
evaluation procedure unchanged. Radius $8$ is again the reported setting,
and Point F1 is recomputed throughout at $\tau=0.05$.

\begin{figure*}[ht!]
\centering
\includegraphics[width=0.88\textwidth]{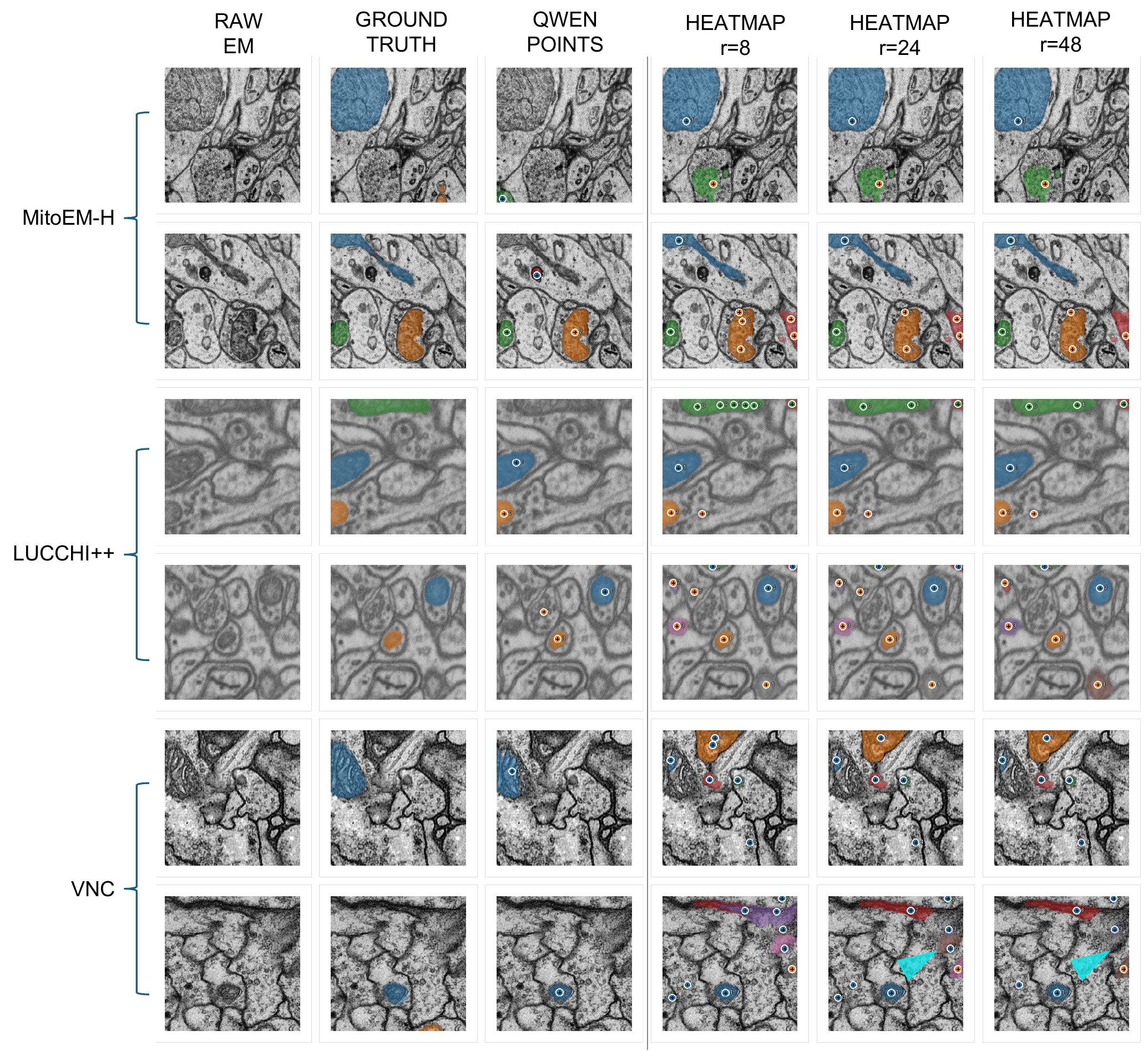}
\caption{\textbf{Qualitative effect of heatmap peak suppression.}
Selected examples show the raw EM image, ground truth, Qwen3-VL SFT point
predictions, and centroid-heatmap points and decoded masks at increasing
peak-NMS minimum distances (radii: r=8, r=24, r=48). Stronger suppression removes nearby heatmap
proposals (cyan arrowheads) while many of the corresponding decoded masks remain similar.}
\label{fig:supp_heatmap_nms_qual}
\end{figure*}

Increasing suppression generally increases Point F1 and reduces
instance-count MAE, whereas AP$_{50}$ changes comparatively little across
the sweep (Fig.~\ref{fig:supp_heatmap_nms}). From radius $8$ to $48$,
AP$_{50}$ changes from $0.888$ to $0.866$ on Lucchi++, from $0.804$ to
$0.779$ on VNC, and from $0.708$ to $0.683$ on MitoEM-H. Over the same
range, count MAE decreases from $3.570$ to $2.374$, from $3.722$ to
$2.418$, and from $1.108$ to $0.991$, respectively. Figure~\ref{fig:supp_heatmap_nms_qual} shows representative examples as
peak suppression increases. Removing nearby heatmap proposals reduces
redundant or unmatched points while many of the decoded object masks remain
similar.

\paragraph{Sensitivity to point placement.}
Distance from the annotated centroid is not by itself a measure of how
useful a point will be to microSAM. We therefore perform a controlled
test on the same $1{,}768$ ground-truth instances using four valid
interior-point strategies: the arithmetic centroid projected inside the
object when necessary, a boundary-near interior point, the
distance-transform maximum, and a random interior point. Each point is
used to prompt the same frozen microSAM backend, so this experiment
isolates the effect of point placement rather than point detection.
Table~\ref{tab:supp_point_placement} reports the resulting masks.

\FloatBarrier
\begin{table}[ht!]
\centering\small
\caption{\textbf{Sensitivity of microSAM to valid point placement.}
Each strategy supplies one point for the same $1{,}768$ ground-truth
instances. Distances are measured in pixels.}
\label{tab:supp_point_placement}
\begin{tabular}{lccccc}
\toprule
Point strategy & Mean selected-mask IoU & IoU$\geq.50$ & IoU$\geq.75$
& Centroid dist. & Boundary dist. \\
\midrule
Projected centroid & 0.828 & 0.954 & 0.833 & 0.21 & 16.80 \\
Boundary-near      & 0.827 & 0.941 & 0.852 & 23.40 & 2.05 \\
Distance maximum   & 0.819 & 0.943 & 0.825 & 9.37 & 20.88 \\
Random interior    & 0.821 & 0.939 & 0.833 & 17.81 & 7.94 \\
\bottomrule
\end{tabular}
\end{table}
\FloatBarrier

Despite large differences in centroid and boundary distance, mean
selected-mask IoU varies only from $0.819$ to $0.828$. Boundary-near points,
for example, lie substantially farther from the arithmetic centroid but
produce nearly the same mean mask IoU as centroid prompts. Centroid-based
point F1 therefore measures agreement with the chosen supervision targets,
but does not completely characterize the usefulness of a valid interior
point to the frozen segmentation backend.

\end{document}